\documentclass[conference]{IEEEtran}
\IEEEoverridecommandlockouts
\usepackage{cite}
\usepackage{amsmath,amssymb,amsfonts}
\usepackage{algorithm}
\usepackage{algorithmic}
\usepackage{graphicx}
\usepackage{textcomp}
\usepackage{xcolor}
\usepackage{times}
\usepackage{amssymb}
\usepackage{amsmath}
\usepackage{url}
\usepackage{subcaption}
\usepackage{float}
\def\BibTeX{{\rm B\kern-.05em{\sc i\kern-.025em b}\kern-.08em
    T\kern-.1667em\lower.7ex\hbox{E}\kern-.125emX}}
    
\usepackage{pgfplots}
\pgfplotsset{compat=1.13}

\definecolor{A}{HTML}{D7191C}
\definecolor{B}{HTML}{FDAE61}
\definecolor{C}{HTML}{ABDDA4}
\definecolor{D}{HTML}{2B83BA}

\newcommand{\etal}{\textit{et al.} }

\newcommand\blfootnote[1]{%
  \begingroup
  \renewcommand\thefootnote{}\footnote{#1}%
  \addtocounter{footnote}{-1}%
  \endgroup
}

\makeatletter
\newcommand\fs@norules{\def\@fs@cfont{\bfseries}\let\@fs@capt\floatc@ruled
  \def\@fs@pre{}%
  \def\@fs@post{}%
  \def\@fs@mid{\kern3pt}%
  \let\@fs@iftopcapt\iftrue}
\makeatother
\floatstyle{norules}
\restylefloat{algorithm}

\begin{document}

\title{t-SNE-CUDA: GPU-Accelerated t-SNE and its Applications to Modern Data}

\author{\IEEEauthorblockN{David M. Chan\textsuperscript{$\dagger$}\IEEEauthorrefmark{1},
Roshan Rao\textsuperscript{$\dagger$}\IEEEauthorrefmark{3}, Forrest Huang\textsuperscript{$\dagger$}\IEEEauthorrefmark{4} and
John F. Canny\IEEEauthorrefmark{5}}
\IEEEauthorblockA{EECS Department,
University of California, Berkeley\\
Berkeley, CA, USA\\
Email: \IEEEauthorrefmark{1}davidchan@berkeley.edu,
\IEEEauthorrefmark{3}roshan\_rao@berkeley.edu,
\IEEEauthorrefmark{4}forrest\_huang@berkeley.edu,
\IEEEauthorrefmark{5}canny@berkeley.edu}}

\maketitle

\begin{abstract} 
Modern datasets and models are notoriously difficult to explore and analyze due to their inherent high dimensionality and massive numbers of samples. Existing visualization methods which employ dimensionality reduction to two or three dimensions are often inefficient and/or ineffective for these datasets. This paper introduces t-SNE-CUDA, a GPU-accelerated implementation of t-distributed Symmetric Neighbour Embedding (t-SNE) for visualizing datasets and models. t-SNE-CUDA significantly outperforms current implementations with 50-700x speedups on the CIFAR-10 and MNIST datasets. These speedups enable, for the first time, visualization of the neural network activations on the entire ImageNet dataset - a feat that was previously computationally intractable. We also demonstrate visualization performance in the NLP domain by visualizing the GloVe embedding vectors. From these visualizations, we can draw interesting conclusions about using the L2 metric in these embedding spaces. t-SNE-CUDA is publicly available at \texttt{\url{https://github.com/CannyLab/tsne-cuda}}.
\end{abstract}

\section{Introduction}

\blfootnote{$\dagger$ Denotes equal contribution among authors}

The recent emergence of large-scale, high-dimensional datasets has been a major factor contributing to advances in the areas of Machine Learning and Artificial Intelligence. While researchers have developed numerous methods for visualizing medium-sized data-sets, such visualizations are often inefficient or ineffective for high-dimensional or large-scale data. This leads to major bottlenecks in a data scientist's research pipeline. Because developing conceptual understandings of the global and local structures of these datasets is vital for successfully developing and improving models, we introduce a fully GPU-based implementation of t-Distributed Stochastic Neighbor Embedding which will allow researchers to explore structure in high-dimensional data efficiently and reduce the burden of forming understandings of the data and models in modern day machine learning tasks.

\begin{figure}
    \centering
    \includegraphics[width=\linewidth]{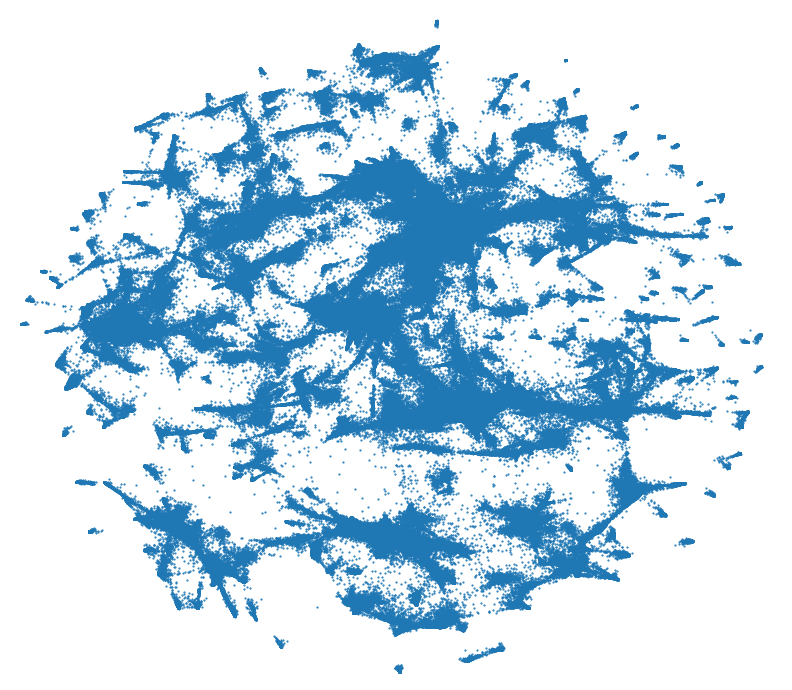}
    \caption{Clustering of the ResNet-200 codes (2048 dimensional) on all 1.2M ImageNet dataset. This embedding was computed in 486s using an NVIDIA Titan X GPU, the same amount of time required to compute the embedding of the MNIST dataset using current state-of-the-art methods.}
    \label{fig:imagenet}
\end{figure}

t-Distributed Stochastic Neighbor Embedding (t-SNE) \cite{maaten2008visualizing} is a dimensionality-reduction method that has recently gained traction in the deep learning community for visualizing model activations and original features of datasets. t-SNE attempts to preserve the local structure of data by matching pairwise similarity distributions in both the higher-dimensional original data space and the lower-dimensional projected space. As opposed to PCA and sub-sampling which both reduce the signal in the data, and hence the quality of the visualization, t-SNE has been shown to generate interesting low-dimensional clusters of data faithful to the distributions in the original data space \cite{maaten2008visualizing}. Unfortunately, current t-SNE implementations are inefficient for visualizing large-scale datasets. All current publicly available implementations executes on the CPU and can require large amounts of time to operate on even modest-sized data (the fastest implementations take over 10 minutes to compute the embedding of the 50,000-image CIFAR-10 dataset), running t-SNE on larger datasets can be intractable. 

In this work, we introduce t-SNE-CUDA, an optimized implementation of the t-SNE algorithm on the GPU. By taking advantage of the natural parallelism in the algorithm, as well as techniques designed for computing the n-body problem, t-SNE-CUDA scales the t-SNE algorithm to large-scale vision datasets such as ImageNet \cite{imagenet_cvpr09}. Our contributions are as follows:

\begin{itemize}
    \item We describe the implementation details of t-SNE-CUDA, our publicly available optimized t-SNE implementation using the Barnes-Hut method and approximate nearest neighbours techniques. t-SNE-CUDA significantly outperforms existing methods with a 50-700x speedup without significantly impacting cluster quality.
    \item We compare and contrast visualizations of real-world large-scale datasets and models, and present some insights into them that can be gleaned from running t-SNE on real-world scale data.
\end{itemize}

\section{Related Work}

t-distributed Symmetric Neighbouring Embedding (t-SNE) \cite{maaten2008visualizing} is widely used in prior work among researchers to visualize data in computer vision and other domains. Van der Maaten \etal qualitatively evaluated t-SNE's performance on both MNIST and CIFAR-10 in \cite{maaten2008visualizing}. DeCAF \cite{donahue2015decaf}, DeVise \cite{frome2013devise} and other tools \cite{izadinia2018viser} all use t-SNE to help understand the activation space of deep convolutional networks. In addition, t-SNE has been used to aid in visualization and understanding of spatio-temporal video data \cite{zhu2017visnavigation, tran2015spatiotemporal}. In many of these works, the analysis was restricted by the efficiency of t-SNE, and thus researches could only analyze subsets of the data or projections of the data into smaller spaces. Our work allows for complete visualizations at the scale required by these papers. 

Current popular implementations of the t-SNE algorithms use tree-based algorithms and approximate nearest neighbours to optimize t-SNE. BH-TSNE \cite{van2014accelerating} and Multicore-TSNE \cite{Ulyanov2016} use the Barnes-hut method to approximate repulsive forces during the training process of t-SNE to reduce computational complexity. Pezzotti \etal \cite{pezzotti2017approximated} use a forest of randomized Kd trees to compute approximate nearest neighbors for the t-SNE algorithm in a steerable manner to emphasize points users deemed important. While the code presented in \cite{pezzotti2017approximated} can be fast, it does not scale to high dimensional data due to the curse of dimensionality and requires very coarse approximations to achieve significant speedups. Section \ref{sec:perf} shows that t-SNE-CUDA clearly outperforms existing publicly available methods by large factors, while maintaining a very high level of accuracy.

In addition to t-SNE, other methods have been explored by data scientists and vision researchers to visualize high dimensional data such as Sammon Mapping \cite{sammon1969nonlinear}, Isomap \cite{tenenbaum2000global}, Locally Linear Embedding \cite{roweis2000nonlinear}, Randomized Principle Component Analysis \cite{rokhlin2009randomized} and Johnson-Lindenstrauss Embedding \cite{larsen2017optimality}. In general, t-SNE has been shown to better preserve local structures and similarity between data points compared to these methods. We redirect interested readers to Van der Maaten's and Arora \etal's works \cite{arora2018analysis,maaten2008visualizing} for a thorough comparison between t-SNE and these visualization methods.

One potential application for fast t-SNE is low-latency, interactive visualization of neural networks. Such visualizations have been shown by \cite{sacha2017you} to increase the productivity of data scientists, and several previous works have explored using t-SNE for such active and interactive visualization. \cite{pezzotti2018deepeyes} suggests using t-SNE for progressive visual analysis of deep neural networks, while \cite{muhlbacher2014opening} suggests t-SNE as a method for increasing user involvement in the training process of DNNs. While we do not explore applications of t-SNE-CUDA to this field of visualization, we believe that it is intriguing and exciting future work, as t-SNE-CUDA is fast enough to visualize training-time embeddings in real-time. 

\section{Methods}
\label{sec:methods}

\subsection{t-SNE}
t-distributed Symmetric Neighbour Embedding (t-SNE) \cite{maaten2008visualizing} is a dimensionality reduction method that reduces high dimensional data into a low dimensional embedding space for primarily visualization applications. t-SNE computes the distribution of pairwise similarities in the high dimensional data space, and attempts to optimize visualization in a low dimensional space by matching the distributions using KL divergence. t-SNE models pairwise similarities between points in both higher and lower dimensional space as conditional probabilities $p_{j|i}$ and $q_{j|i}$. The conditional probability $p_{j|i}$, for instance, can be interpreted as the probability that a point $j$ is a neighbor of point $i$ in the higher dimensional space. t-SNE models the probabilities as a Gaussian distribution around each data points in the higher dimensional space,

\begin{equation}
   p_{j|i} = \frac{\exp(-||x_i-x_j||^2/2\sigma_i^2)}{\sum_{k \neq i}\exp(-||x_i-x_k||^2/2\sigma_i^2)} 
   \label{eq:p_j_given_i}
\end{equation}
and models the target distribution of pairwise similarities in the lower dimensional embedding space using a Student's t-distribution around each data point to overcome the overcrowding problem in the Gaussian distribution:
\begin{equation}
    q_{ij} = \frac{(1 + ||y_i - y_j||^2)^{-1}}{\sum_{k \neq l} (1 + ||y_k - y_l||^2)^{-1}}
    \label{eq:qij}
\end{equation}
t-SNE then minimizes the KL divergence between the distributions, which conserves the local structure of data points across the higher and lower dimensional spaces. 

To minimize the KL divergence, t-SNE employs gradient descent with the gradient computed as follows:
\begin{equation}
    \frac{\partial C}{\partial y_i} = 4 \sum_j (p_{ij}-q_{ij})q_{ij}(\mathbf{y}_i - \mathbf{y}_j)
    \label{eq:tsne_gradient}
\end{equation}

\begin{equation}
    Z = \sum_{k \neq l}(1+||y_k-y_l||^2)^{-1}
    \label{eq:tsne_normalization}
\end{equation}
with $p_{ij}$(and similarly, $q_{ij}$) computed as the symmetrized joint probabilities of $p_{i|j}$ and $p_{j|i}$ such that $p_{ij} = \dfrac{p_{i|j} + p_{j|i}}{2N}$.

The t-SNE gradient computation can be reformulated as an N-body simulation problem by rearranging the terms into attractive forces and repulsive forces:
\begin{equation}
    F_{attr} = \sum_{j \in [1, \ldots, N], j \neq i}  p_{ij}q_{ij}Z(\mathbf{y}_i - \mathbf{y}_j)
    \label{eq:attr_force}
\end{equation}

\begin{equation}
    F_{rep} = -\sum_{j \in [1, \ldots, N], j \neq i} q_{ij}^2Z(\mathbf{y}_i - \mathbf{y}_j)
    \label{eq:rep_force}
\end{equation}

\begin{equation}
    \frac{\partial C}{\partial y_i} = 4 (F_{attr} + F_{rep})
    \label{eq:bh_tsne_gradient}
\end{equation}

\subsection{Attractive Forces}
For the attractive forces, the term $p_{ij}$ in \eqref{eq:attr_force} decreases exponentially with squared distance between the points $x_i, x_j$ in the higher dimensional space. If we take $p_{ij} = 0$ beyond some threshold distance, this introduces significant sparsity into the attractive force calculation. In practice, instead of defining a threshold distance, the K nearest neighbors of each point are obtained and $p_{j|i}$ for points outside the K nearest neighbors of $i$ is taken to be zero. As symmetrizing $p_{j|i}$ can at most double the number of nonzero values, $p_{ij}$ can be stored as a sparse matrix with at most $N * 2K$ nonzero entries instead of $N^2$ nonzero entries. By iterating over only nonzero values of $p_{ij}$, the attractive force computation becomes linear in $N$. Empirically, using relatively small values of $K$ (between 32 and 150) are sufficient to achieve reasonable visualizations.

There are a number of methods for computing the k-Nearest Neighbors of a point. The most common implementation is the ``Kd Tree" which partitions the search space using the nodes of a tree. While the Kd tree is able to find all nearest neighbors in $O(dN\log(N))$ time (where $d$ is the number of dimensions), this is only sufficient for low dimensional queries. It is computationally expensive for higher dimensions, which are common in many modern data-analysis problems as the dimension dominates the computational cost. In general because $d >> K$, we end up finding the exact nearest neighbors in $\approx O(KN)$ time which is no better than a naive search, whereas better performance can be achieved for finding \emph{approximate} nearest neighbors.

To solve the approximate nearest neighbor problem, Pezzotti \etal \cite{pezzotti2017approximated} use a random forest of approximate K-d trees, which allowed them to compute the nearest neighbors in a faster manner, however their approach still suffers from high dimension. There are, however, a number of modern techniques for approximate neighbor selection in high dimension that further reduces the computational complexity. t-SNE-CUDA uses the FAISS \cite{JDH17} library, which provides an efficient, easy to use GPU implementation of similarity search. FAISS is designed for ``Billion scale data," and allows us to scale our library to very large datasets.

FAISS is based on locally-sensitive hashing around Voronoi cells in the data. It uses the IDFVAC indexing structure presented in \cite{jegou2011product} as an indexing structure. Database vectors $y$ are encoded using \begin{equation}y \approx q(y) = q_1(y_ + q2(y - q_1(y))\end{equation} where $q_1$ and $q_2$ are quantizing functions. The $q_1$ function is a coarse quantizer, while the $q_2$ quantizer is a more fine approximation encoding the residual value. We then rephrase the nearest neighbor problem \begin{equation}\mathcal{N}_x = k\text{-argmin}_{y \in N} ||x - y_i||\end{equation} as an approximate asymmetric distance problem. First, the algorithm compute \begin{equation}\mathcal{N}^{IVF}_x = \tau\text{-argmin}_{c \in \mathcal{C}} ||x - c||\end{equation} This gives a coarse grained approximation of the location of the point $x$ in terms of ``centroids" in $\mathcal{C}$. We then construct our nearest neighbors \begin{equation}\mathcal{N_x} \approx k\text{-argmin}_{y \in N | q_1(y) \in \mathcal{N}^{IVF}_x} ||x - q(y)||\end{equation} By storing the index as an inverted file, and grouping the vectors around the centroids, we can achieve a look-up by linearly scanning $O(\tau)$ inverted lists. 

For our implementation, we choose $|C| = \sqrt{N}$, and train the vectors $C$ using the k-Means algorithm. Thus, $q_1$ is the id of the nearest centroid. $q_2$ is much more precise, and is selected using product quantization \cite{jegou2011product} which interprets the vector $y$ as a set of quantized sub-vectors. For more details of this algorithm, we refer interested readers to \cite{JDH17}. The parameter $\tau$, selected by the user, controls the accuracy of the KNN algorithm. 

Once the k-Nearest Neighbors are computed, we can compute a sparse matrix $P_{ij}$ which stores the nonzero values of $p_{ij}$. The attractive force can then be computed efficiently by decomposing it as a series of matrix operations. Let $Q_{ij}$ represent the matrix of $q_{ij}$ values, and $Y$ represent the $N \times 2$ matrix of points in the lower dimensional space. Additionally, let $O$ be a $N \times 2$ matrix of ones and $\odot$ represent the Hadamard product of two matrices. We first distribute the multiplication of $F_{attr}$ giving,
\begin{equation}
    F_{attr} = 4Ny_i\sum_j p_{ij}q_{ij} - 4N\sum_j p_{ij}q_{ij}y_j
\end{equation}
\begin{equation}
    F_{attr} = 4N((P_{ij} \odot Q_{ij})O \odot Y - (P_{ij} \odot Q_{ij})Y)
\end{equation}
Since $P \odot Q$ is computed only once, this becomes one matrix-matrix subtraction, two Hadamard products, and two matrix-matrix multiplications. To achieve the $O(NK)$ run-time, we represent $P_{ij}$ as a sparse matrix with a nonzero value at $i, j$ iff $j$ is a neighbor of $i$. $Q_{ij}$ is never computed in its entirety; instead, the matrix $P \odot Q$ is computed directly by iterating over nonzero values of $P$. The matrix-matrix multiplications are performed using cuSPARSE.

\subsection{Repulsive Forces}
The repulsive force is more challenging to approximate because the long tails of the Student's T-distribution create reasonably strong repulsive forces even at intermediate distances. This is where the Barnes-Hut approximation occurs. At each iteration, the lower dimensional points $y_1, \ldots, y_N$ are placed in a quad tree. Then, for each point a depth-first-search is performed on the quad tree. When looking at a quad tree cell centered at $y_{cell}$ with radius $r_{cell}$, the following condition is evaluated:
\begin{equation}
    \frac{r_{cell}}{||y_i - y_{cell}||} < \theta
\end{equation}
\label{eq:recurse_condition}
If this evaluates to true, then the cell is deemed far enough away to be used as a summary of the forces for all children and the recursion halts. $\theta$ is a parameter that controls the accuracy of the approximation with $\theta = 0$ giving the $O(N^2)$ algorithm. If a cell is deemed far enough away, the force on point $y_i$ is given by
\begin{equation}
    \frac{N_{cell}(y_i - y_{cell})}{(1 + ||y_i - y_{cell}||^2)^2}\approx \sum_{j \in cell} q_{ij}^2Z^2(y_i - y_j)
\end{equation}
Note that unlike in \eqref{eq:rep_force}, here the normalization constant $Z$ is squared. However, the normalization constant can also be approximated by simultaneously computing an approximate reduction over $q_{ij}$. 

Given this formula, the approximation is performed in 5 steps: 1) Compute a bounding box of the points, 2) Build a hierarchical decomposition by inserting points into a quad tree, 3) Compute the number of points in each internal cell, 4) Sort points by spatial distance, 5) Compute forces on points using the quad tree.

The code that t-SNE-CUDA uses for computing these steps is adapted to fit the t-SNE objective from \cite{Burtscher2011}. \cite{Burtscher2011} provides an implementation of GPU tree construction and traversal that attempts to minimize thread divergence, wait times, and other sources of slowdowns on the GPU. 

\subsection{Algorithm}

Algorithm \ref{alg:basic} gives the full outline of the discussed sections. Our full implementation is publicly available, so we omit many of the code details, and reserve space in the paper for a mathematical overview of the algorithm, and a discussion of the performance. 

\begin{algorithm}[!tb]
 \caption{General T-SNE-CUDA Algorithm\label{alg:basic}}
 \begin{algorithmic}[1]
 \renewcommand{\algorithmicrequire}{\textbf{Input:} $N \times d-$dimensional array of data}
 \renewcommand{\algorithmicensure}{\textbf{Output:} $N \times 2-$dimensional projection}
 \REQUIRE
 \ENSURE
  \STATE FAISS Computation (approximate k-NN)
  \STATE Use pairwise distances to compute sparse matrix of $P_{ij}$
  \FOR {$i = 1$ to $\bf{convergence}$}
  \STATE R-Force Tree Building (build tree for Barnes-Hut)
  \STATE R-Force Computation (use tree to compute approximate repulsive forces)
  \STATE Compute $P_{ij} \odot Q_{ij}$
  \STATE A-Force cuSPARSE (sparse matrix times dense vector)
  \STATE Apply Forces (apply forces to points in lower dimensional space)
  \ENDFOR
 \RETURN lower dimensional projection
 \end{algorithmic} 
 \end{algorithm}

\section{Performance}
\label{sec:perf}

In this section we discuss the performance of our algorithm through the lens of some real-world empirical experiments. 

\subsection{Experiments}

\subsubsection{Target Environment}

We perform experiments given in this paper using a system with an Intel i7-5820K Processor, containing 6 physical cores (12 with hyper threading) and 64GB of DDR4 RAM. The GPU in use on this system is the NVIDIA Titan-X Maxwell edition GPU, with 3072 CUDA cores clocked at 1.0 GHz and 12GB of GDDR5 memory. It supports a maximum memory speed of 7.0Gbps, with a maximum memory bandwidth of 336.5Gbps. The GM200 chip (Titan-X) has 3072Kb of L2 cache, and 24 Streaming Multiprocessors (6GPCs), with a theoretical peak performance of 6.12TFLOPs for single precision floating point operations. The CPU platform has a theoretical peak performance of 691.2GFlops, giving a peak-to-peak theoretical margin of 8.85x. The CUDA grid sizes have been optimized according to our unique GPU, and for each kernel using a grid search across a set of synthetic problems - For brevity, we provide the full details of implementation as well as optimized grid sizes in our online repository.

\subsubsection{Datasets}

\textbf{Simulated Data:} It is important to be able to benchmark methods in a controlled experiment, so we construct simulated data consisting of equal-sized clusters of points sampled from four high-dimensional Gaussian distributions. In these experiments, we are able to vary the size and dimensions of the data, and effectively examine the performance of different algorithms in a controlled environment. 

\noindent\textbf{MNIST:} The MNIST dataset \cite{mnist} is a classic computer vision dataset consisting of 60,000 training images, and 10,000 testing images depicting different handwritten digits (numerals 0-9). Each of these digits is black and white, with dimensions 28x28 constituting a 784 dimensional image space. 

\noindent\textbf{CIFAR:} The CIFAR datasets \cite{cifar} both consist of a 50,000 image subset from the larger tiny-image dataset. The datasets have images of 10 (Resp. 100) classes such as "ship", "car", "horse", "frog" etc. The images from the CIFAR-10/100 datasets are full color images with dimensions of 32x32x3, giving a 3072 dimensional data space to explore.

\subsection{Synthetic Data}
\label{sec:synth}

\begin{figure*}[!htb]
    \centering
    \begin{tikzpicture}
    \begin{axis}
    [
        width=\linewidth,
        height=0.35\linewidth,
        xlabel=Number of Points (Thousands),
        ylabel=Time (s),
        xmode=log,
        tick label style={font=\footnotesize},
        log ticks with fixed point,
         log basis x={10},
        ymode=log,
         legend pos=north west
    ]
        \addplot
        plot[blue, mark options={blue}] coordinates {
            (0.5, 8.15)
            (1,17.78)
            (2,39.43)
            (4,85.74)
            (8,173.53)
            (16,351.47)
            (32,809.46)
            (64,2119.68)
            (128,6734.36)
            (512,78461.80)
            
        };
        \addplot
        plot[blue,  mark=none, dashed, forget plot] coordinates {
            (512,78461.80)
            (5000,896282)

        };
        \addplot
        plot[red, mark options={red}] coordinates {
            (0.5,2.57)
            (1,6.38)
            (2,18.49)
            (4,33.38)
            (8,123.72)
            (16,340.75)
            (32,1011.03)
            (64,2521.78)
            (128,8015.17)
            (512,42791.82)
        };
         \addplot
        plot[red,  mark=none, dashed, forget plot] coordinates {
            (512,42791.82)
            (5000,492513.25)

        };
        \addplot
        plot[orange, mark options={orange}] coordinates {
            (0.5,2.74)
            (1,5.00)
            (2,11.07)
            (4,31.67)
            (8,198.15)
            (16,198.15)
            (32,473.75)
            (64,1759.41)
            (128,4716.18)
            (512,56063.06)
        };
        \addplot
        plot[orange,  mark=none, dashed, forget plot] coordinates {
            (512,56063.06)
            (5000,640575)

        };
        \addplot
        plot[purple, mark options={purple}] coordinates {
            (0.5,1.32)
            (1,2.52)
            (2,5.67)
            (4,16.47)
            (8,41.28)
            (16,87.53)
            (32,201.04)
            (64,529.56)
            (128,2251.68)
            (512,18503.79)
        };
        \addplot
        plot[purple,  mark=none, dashed, forget plot] coordinates {
            (512,18503.79)
            (5000,212135.148)

        };
        \addplot
        plot[green, mark options={green}] coordinates {
            (0.5,0.84)
            (1,0.90)
            (2,1.019)
            (4,1.133)
            (8,1.556)
            (16,1.792)
            (32,2.334)
            (64,3.695)
            (128,6.553)
            (512,40.341)
            (5000,987)
        };
        \legend{SkLearn, MULTICORE-1, MULTICORE-4, BH-TSNE, t-SNE-CUDA (Ours)}
    \end{axis}        
\end{tikzpicture}
    \caption{Time taken compared to other state-of-the-art algorithms on synthetic datasets with 50 dimensions and four clusters for varying numbers of points. Note the log scale on both the $x$ and time axis, and that the scale of the $x$ axis is in thousands of points (thus, the values on the $x$ axis range from 0.5K to 10M points). Dashed lines represent projected times. Projected scaling assumes an $O(n\log n)$ implementation. For small numbers of points, the GPU is not fully saturated leading to better than $O(n\log n)$ scaling. When the GPU becomes fully engaged, our algorithm exhibits a clear $O(n\log n)$ scaling pattern.}
    \label{fig:synthetic_comparison}
\end{figure*}
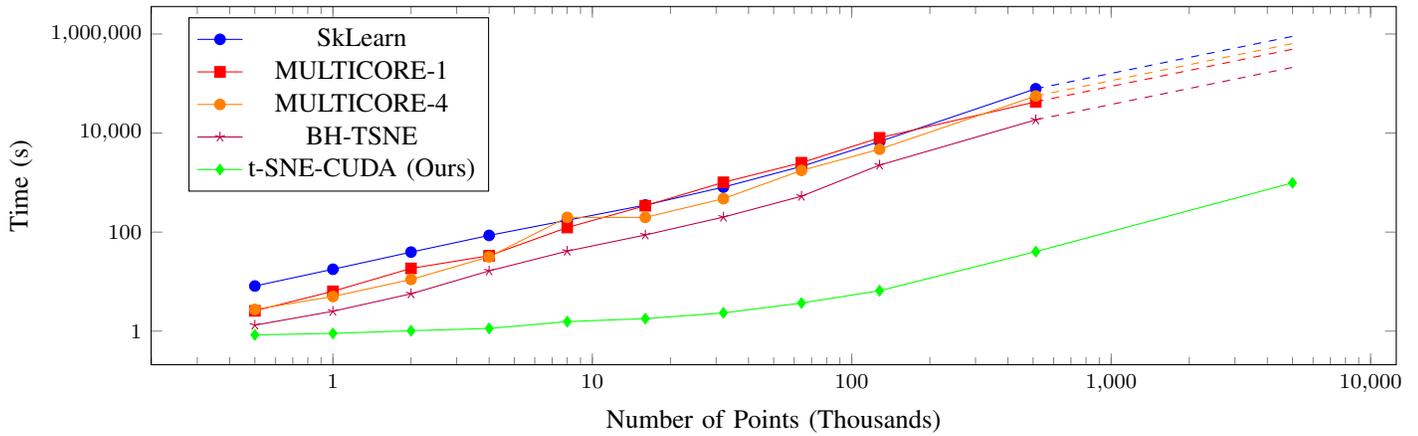

Figure \ref{fig:synthetic_comparison} compares the running time of our algorithm with existing implementations on a synthetic dataset that consists of various number of Gaussian-distributed data points for 50 dimensions in 4 clusters. We can see from the Figure that in general we are one to two orders of magnitude faster than even the best current implementations. At 32,000 points, we achieve a 346x speedup over SkLearn with 50 dimensions, and a 86x speedup over Multicore t-SNE. At 512,000 points we achieve a speedup of 1946.92x over SkLearn and a 459.13x speedup over Multicore t-SNE. We also experimented with six and eight core variants of the Multicore t-SNE algorithm, however both had worse or similar performance to the four core variant (MULTICORE-4) of the algorithm.

\subsection{Real-World Datasets}

\begin{figure}
\centering
\begin{tikzpicture}
\begin{axis}[
    xbar stacked,
    legend style={
    legend columns=4,
        at={(xticklabel cs:0.5)},
        anchor=north,
        draw=none
    },
    ytick=data,
    axis y line*=none,
    axis x line*=bottom,
    tick label style={font=\footnotesize},
    legend style={font=\footnotesize},
    label style={font=\footnotesize},
    xtick={0,1000,...,5000},
    width=.7\linewidth,
    bar width=2mm,
    yticklabels={
    {SkLearn}, 
    {MULTICORE-1}, 
    {BH-TSNE}, 
    {MULTICORE-4}, 
    {t-SNE-CUDA (Ours)}},
    xmin=0,
    xmax=5000,
    area legend,
    y=4mm,
    enlarge y limits={abs=0.625},
    nodes near coords,
    nodes near coords style={text=black, at ={(\pgfplotspointmeta,\pgfplotspointy)},anchor=west},
    visualization depends on=y \as \pgfplotspointy,
    every axis plot/.append style={fill}
]
\addplot[A, nodes near coords=\pgfmathprintnumber{\pgfplotspointmeta} (702.2x), nodes near coords style={/pgf/number format/.cd,fixed zerofill,precision=2}] coordinates
  {(3665.44 ,0) (0,1) (0,2) (0,3) (0,4)};
\addplot[B, nodes near coords=\pgfmathprintnumber{\pgfplotspointmeta} (115.6x), nodes near coords style={/pgf/number format/.cd,fixed zerofill,precision=2}] coordinates
  {(0,0) (603.46,1) (0,2) (0,3) (0,4)};
\addplot[C, nodes near coords=\pgfmathprintnumber{\pgfplotspointmeta} (166.8x), nodes near coords style={/pgf/number format/.cd,fixed zerofill,precision=2}] coordinates
  {(0,0) (0,1) (870.71,2) (0,3) (0,4)};
\addplot[D, nodes near coords=\pgfmathprintnumber{\pgfplotspointmeta} (52.8x), nodes near coords style={/pgf/number format/.cd,fixed zerofill,precision=2}] coordinates
  {(0,0) (0,1) (0,2) (275.46,3) (0,4)};
\addplot[blue, nodes near coords=\pgfmathprintnumber{\pgfplotspointmeta} (1.0x), nodes near coords style={/pgf/number format/.cd,fixed zerofill,precision=2}] coordinates
  {(0,0) (0,1) (0,2) (0,3) (5.22,4)};
\end{axis}  

\end{tikzpicture}
\caption{The performance of t-SNE-CUDA compared to other state-of-the-art implementations on the CIFAR-10 dataset. t-SNE-CUDA runs on the output of a classifier on the CIFAR-10 training set (50000 images x 1024 dimensions) in under 6 seconds. While we can run on the full pixel set in under 12 seconds, Euclidean distance is a poor metric in raw pixel space leading to poor quality embeddings. }
\label{fig:cifar}
\end{figure}
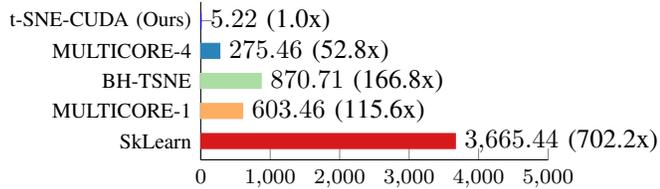

Figure \ref{fig:mnist} and \ref{fig:cifar} shows the time taken by our algorithm to compute the embeddings for the MNIST and CIFAR datasets and the speedup compared to current state-of-the-art CPU implementations. t-SNE-CUDA significantly outperforms the popular SKLearn toolkit with more than 700 times speedup over the CIFAR-10 dataset, and with more than 650 times speedup over the MNIST dataset. t-SNE-CUDA also achieves more than 50 times speedup over the state-of-the-art implementations in both datasets.

\begin{figure}
\centering
\begin{tikzpicture}
\begin{axis}[
    width=\linewidth,
    height=0.35\linewidth,
    xbar stacked,
    legend style={
    legend columns=4,
        at={(xticklabel cs:0.5)},
        anchor=north,
        draw=none
    },
    ytick=data,
    axis y line*=none,
    axis x line*=bottom,
    tick label style={font=\footnotesize},
    legend style={font=\footnotesize},
    label style={font=\footnotesize},
    xtick={0,1000,...,5000},
    width=.6\linewidth,
    bar width=2mm,
    yticklabels={
    {SkLearn}, 
    {MULTICORE-1}, 
    {BH-TSNE}, 
    {MULTICORE-4}, 
    {t-SNE-CUDA (Ours)}},
    xmin=0,
    xmax=4600,
    area legend,
    y=4mm,
    enlarge y limits={abs=0.625},
    nodes near coords,
    nodes near coords style={text=black, at ={(\pgfplotspointmeta,\pgfplotspointy)},anchor=west},
    visualization depends on=y \as \pgfplotspointy,
    every axis plot/.append style={fill}
]
\addplot[A, nodes near coords=\pgfmathprintnumber{\pgfplotspointmeta} (652.8x), nodes near coords style={/pgf/number format/.cd,fixed zerofill,precision=2}] coordinates
  {(4556.58 ,0) (0,1) (0,2) (0,3) (0,4)};
\addplot[B, nodes near coords=\pgfmathprintnumber{\pgfplotspointmeta} (190.1x), nodes near coords style={/pgf/number format/.cd,fixed zerofill,precision=2}] coordinates
  {(0,0) (1327.07,1) (0,2) (0,3) (0,4)};
\addplot[C, nodes near coords=\pgfmathprintnumber{\pgfplotspointmeta} (165.7x), nodes near coords style={/pgf/number format/.cd,fixed zerofill,precision=2}] coordinates
  {(0,0) (0,1) (1156.70,2) (0,3) (0,4)};
\addplot[D, nodes near coords=\pgfmathprintnumber{\pgfplotspointmeta} (71.8x), nodes near coords style={/pgf/number format/.cd,fixed zerofill,precision=2}] coordinates
  {(0,0) (0,1) (0,2) (501.41,3) (0,4)};
\addplot[blue, nodes near coords=\pgfmathprintnumber{\pgfplotspointmeta} (1.0x), nodes near coords style={/pgf/number format/.cd,fixed zerofill,precision=2}] coordinates
  {(0,0) (0,1) (0,2) (0,3) (6.98,4)};
\end{axis}  

\end{tikzpicture}
\caption{The performance of t-SNE-CUDA compared to other state-of-the-art implementations on the MNIST dataset. t-SNE-CUDA runs on the raw pixels of the MNIST dataset (60000 images x 768 dimensions) in under 7 seconds.}
\label{fig:mnist}
\end{figure}
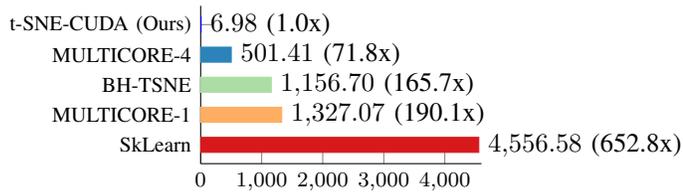

\begin{figure}[!htb]
    \includegraphics[width=\linewidth]{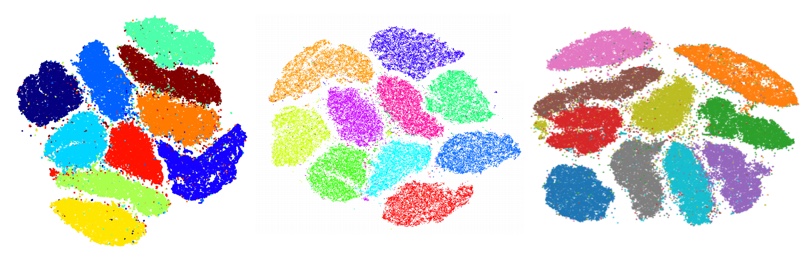}
    \caption{Comparison of different clustering techniques on the MNIST dataset in pixel space. Left: MULTICORE-4 (501s), Middle: BH-TSNE (1156s), Right: t-SNE-CUDA (Ours, 6.98s)}
    \label{fig:vs_fig_mnist}
\end{figure}
\begin{figure}[!htb]
    \includegraphics[width=\linewidth]{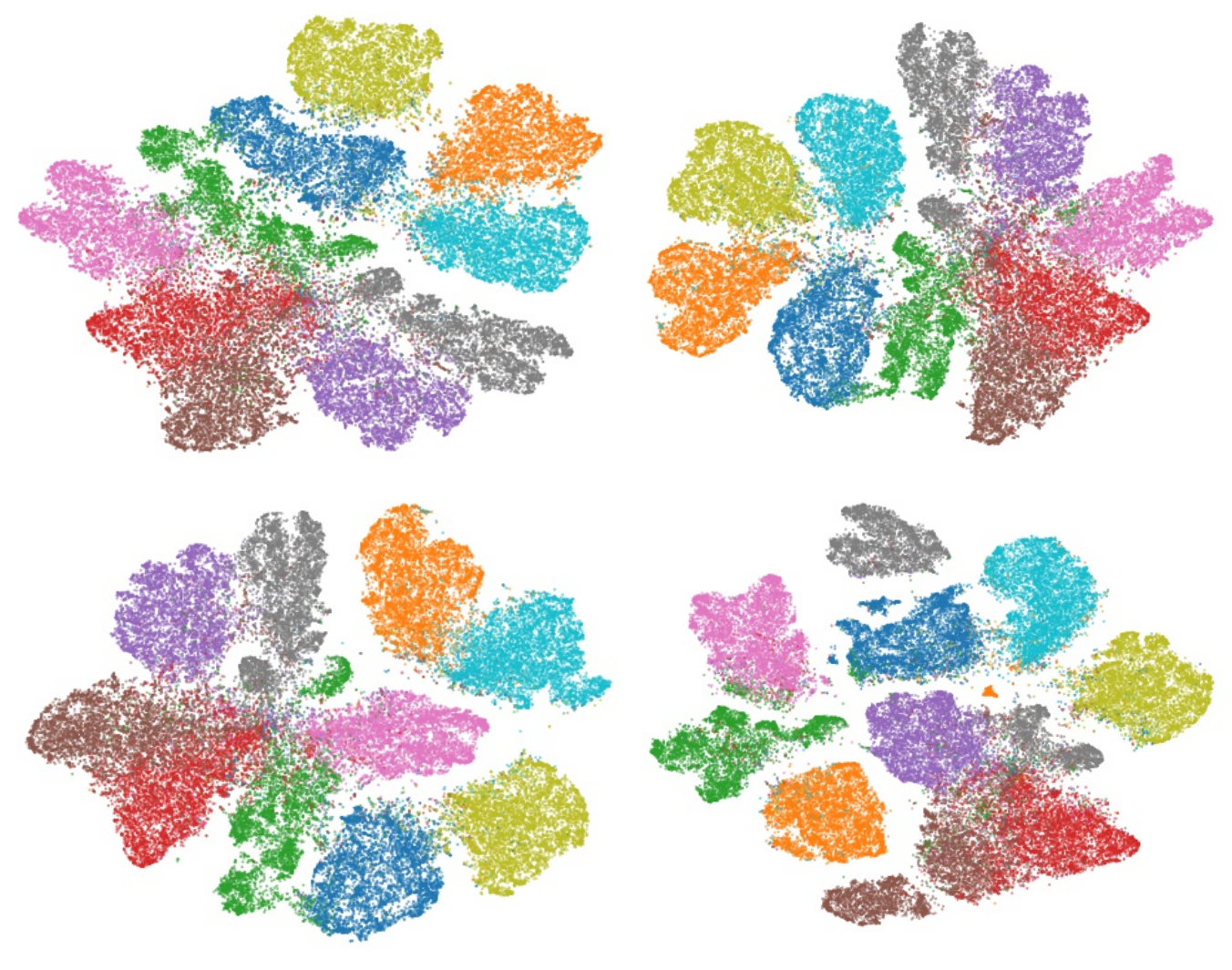}
    \caption{Comparison of clustering techniuqes on the LeNet \cite{lecun1998gradient} codes on the CIFAR-10 dataset. Top-left: SkLearn (3665.44s), Top-right: BH-TSNE (870.71s), Bottom-left: MULTICORE-4(275.46s), Bottom-right: t-SNE-CUDA (Ours, 5.22s).}
    \label{fig:vs_fig_cifar}
\end{figure}

\newpage
\phantom{ }
\newpage

Moreover, the quality of the embeddings produced by t-SNE-CUDA do not differ significantly from the state-of-the-art implementations. Figures  \ref{fig:vs_fig_mnist} and \ref{fig:vs_fig_cifar} show that t-SNE-CUDA does not compromise on the quality of the clusters while significantly outperforming other state-of-the-art methods in terms of speed.

\subsection{Kernel Performance}

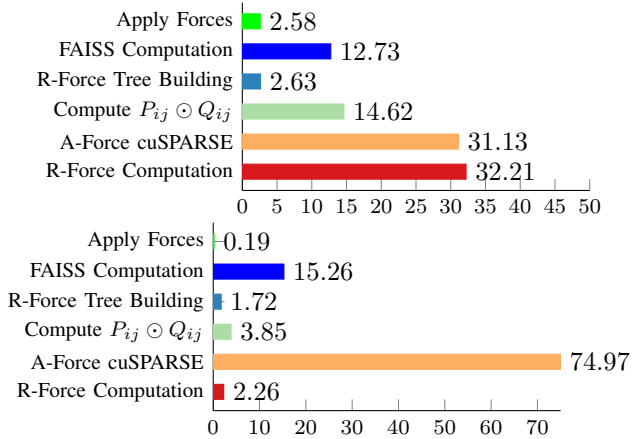
\begin{figure}
\centering
\begin{tikzpicture}
\begin{axis}[
    xbar stacked,
    legend style={
    legend columns=4,
        at={(xticklabel cs:0.5)},
        anchor=north,
        draw=none
    },
    ytick=data,
    axis y line*=none,
    axis x line*=bottom,
    tick label style={font=\footnotesize},
    legend style={font=\footnotesize},
    label style={font=\footnotesize},
    xtick={0,5,...,50},
    width=.7\linewidth,
    bar width=2mm,
    yticklabels={
    {R-Force Computation}, 
    {A-Force cuSPARSE}, 
    {Compute $P_{ij} \odot Q_{ij}$}, 
    {R-Force Tree Building}, 
    {FAISS Computation},
    {Apply Forces}},
    xmin=0,
    xmax=50,
    area legend,
    y=4mm,
    enlarge y limits={abs=0.625},
    nodes near coords,
    nodes near coords style={text=black, at ={(\pgfplotspointmeta,\pgfplotspointy)},anchor=west},
    visualization depends on=y \as \pgfplotspointy,
    every axis plot/.append style={fill}
]
\addplot[A] coordinates
  {(32.21 ,0) (0,1) (0,2) (0,3) (0,4) (0,5)};
\addplot[B] coordinates
  {(0,0) (31.13,1) (0,2) (0,3) (0,4) (0,5)};
\addplot[C] coordinates
  {(0,0) (0,1) (14.62,2) (0,3) (0,4) (0,5)};
\addplot[D] coordinates
  {(0,0) (0,1) (0,2) (2.63,3) (0,4) (0,5)};
\addplot[blue] coordinates
  {(0,0) (0,1) (0,2) (0,3) (12.73,4) (0,5)};
\addplot[green] coordinates
  {(0,0) (0,1) (0,2) (0,3) (0,4) (2.58,5)};
\end{axis}  
\end{tikzpicture}
\begin{tikzpicture}
\begin{axis}[
    xbar stacked,
    legend style={
    legend columns=4,
        at={(xticklabel cs:0.5)},
        anchor=north,
        draw=none
    },
    ytick=data,
    axis y line*=none,
    axis x line*=bottom,
    tick label style={font=\footnotesize},
    legend style={font=\footnotesize},
    label style={font=\footnotesize},
    xtick={0,10,...,75},
    width=.7\linewidth,
    bar width=2mm,
    yticklabels={
    {R-Force Computation}, 
    {A-Force cuSPARSE}, 
    {Compute $P_{ij} \odot Q_{ij}$}, 
    {R-Force Tree Building}, 
    {FAISS Computation},
    {Apply Forces}},
    xmin=0,
    xmax=75,
    area legend,
    y=4mm,
    enlarge y limits={abs=0.625},
    nodes near coords,
    nodes near coords style={text=black, at ={(\pgfplotspointmeta,\pgfplotspointy)},anchor=west},
    visualization depends on=y \as \pgfplotspointy,
    every axis plot/.append style={fill}
]
\addplot[A] coordinates
  {(2.26 ,0) (0,1) (0,2) (0,3) (0,4) (0,5)};
\addplot[B] coordinates
  {(0,0) (74.97,1) (0,2) (0,3) (0,4) (0,5)};
\addplot[C] coordinates
  {(0,0) (0,1) (3.85,2) (0,3) (0,4) (0,5)};
\addplot[D] coordinates
  {(0,0) (0,1) (0,2) (1.72,3) (0,4) (0,5)};
\addplot[blue] coordinates
  {(0,0) (0,1) (0,2) (0,3) (15.26,4) (0,5)};
\addplot[green] coordinates
  {(0,0) (0,1) (0,2) (0,3) (0,4) (0.185,5)};
\end{axis}  
\end{tikzpicture}
\caption{Percentage of the time spent in each of the top-6 most expensive kernels. The top graph shows these values computed for 500,000 synthetic points, while the bottom graph shows the same computation for 5M points. In both cases the points are sampled from 4 Gaussian distributions of 50 dimensions.}
\label{fig:kerneltimes}
\end{figure}

Figure \ref{fig:kerneltimes} gives a general breakdown of the performance of t-SNE-CUDA by percentage of time taken in each part of the algorithm for 500,000 points and 5M synthetic points. There are two phases to t-SNE-CUDA as discussed in Section \ref{sec:methods}, however clearly as the number of points grows larger, the second phase (computing the attractive and repulsive forces) dominates the construction of the nearest neighbors. It is interesting to note that with an increased dataset size, the repulsive force computation time significantly decreased in terms of percentage. The reason for this is that the attractive forces are computed using a sparse matrix multiplication as part of cuBLAS, and the run-time of the Barnes-Hut part of the force computation is dominated by the sparse matrix multiply operation. Future work could improve the sparse matrix multiply operation to further improve performance. 

Moreover, cuSPARSE calls dominate the running time for large numbers of points. This is likely due to the fact that the computation of attractive forces breaks into a large sparse matrix dense vector multiplication. Because the sparsity pattern is not well organized (the organization of the sparsity depends on the clustering), it makes it difficult, and rather expensive to compute these values.

Because our kernels are mostly performing integer operations, we achieve a very small amount of the peak theoretical performance of the GPU. Since many of our kernels are memory/offset computations and transforms, we perform very little floating point work, and thus almost all of our kernels achieve $\approx$ 10\% of the peak performance of the GPU. The reason that floating point operations are not performed is due to the sparse matrix multiply operation in the attractive force computation kernel. During this sparse multiplication, a large amount of time is spent computing offsets for the different indices, instead of actually computing multiplications. Because this offset-computation performance requires integer arithmetic (which the GPU is not optimized for), we have decreased performance (while performing close to the roof-line for integer computation). Our more heavily floating point focused kernels, such as the integration kernel, can achieve better throughput; For 5M points, we achieve 68.23\% of peak performance, close to the roof-line for memory operations. Indeed, our kernels are instead generally memory latency bounded - almost 68.34\% of the stalls are caused by memory dependencies. One clear direction for future work is to further examine the memory usage of the project, and to investigate whether better algorithms could be devised to reduce the number of memory stalls.  

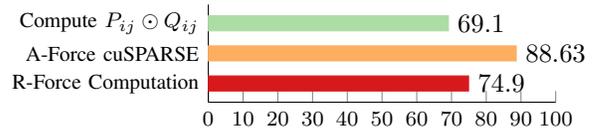
\begin{figure}
\centering
\begin{tikzpicture}
\begin{axis}[
    xbar stacked,
    legend style={
    legend columns=4,
        at={(xticklabel cs:0.5)},
        anchor=north,
        draw=none
    },
    ytick=data,
    axis y line*=none,
    axis x line*=bottom,
    tick label style={font=\footnotesize},
    legend style={font=\footnotesize},
    label style={font=\footnotesize},
    xtick={0,10,...,100},
    width=.7\linewidth,
    bar width=2mm,
    yticklabels={
    {R-Force Computation}, 
    {A-Force cuSPARSE}, 
    {Compute $P_{ij} \odot Q_{ij}$}, 
    {R-Force Tree Building}, 
    {FAISS Computation},
    {Apply Forces}},
    xmin=0,
    xmax=100,
    area legend,
    y=4mm,
    enlarge y limits={abs=0.625},
    nodes near coords,
    nodes near coords style={text=black, at ={(\pgfplotspointmeta,\pgfplotspointy)},anchor=west},
    visualization depends on=y \as \pgfplotspointy,
    every axis plot/.append style={fill}
]
\addplot[A] coordinates
  {(74.9,0) (0,1) (0,2)};
\addplot[B] coordinates
  {(0,0) (88.63,1) (0,2)};
\addplot[C] coordinates
  {(0,0) (0,1) (69.1,2)};
\end{axis}  
\end{tikzpicture}
\caption{Percentage of kernel occupancy on the GPU for the top-3 kernels in a run of 500,000 Gaussian symmetric points with four clusters.}
\label{fig:occupancy}
\end{figure}

In general, the kernels that dominate the running time are not troubled by occupancy, and instead are bounded by the sheer number of memory operations (along with the integer arithmetic). Figure \ref{fig:occupancy} shows the occupancy of each of the kernels. By examining the occupancy we can see that while warps may be stalling, we are generally able to have many threads resident at the same time which improves performance. In addition, we find that our average SM utilization is high at above 90\% on average for all of the kernels, meaning that we are fully taking advantage of the GPU resources on our device. 

\section{Exploring Data with t-SNE-CUDA}

In this section, we analyze some data visualizations that are made possible with t-SNE-CUDA's improved performance. By leveraging the power to do pixel-level exploration on medium-sized data, and code-level exploration on very large-scale datasets, we can draw interesting conclusions about popular machine learning datasets.

\subsection{Why is CIFAR harder than MNIST?}

\begin{figure}
    \includegraphics[width=\linewidth]{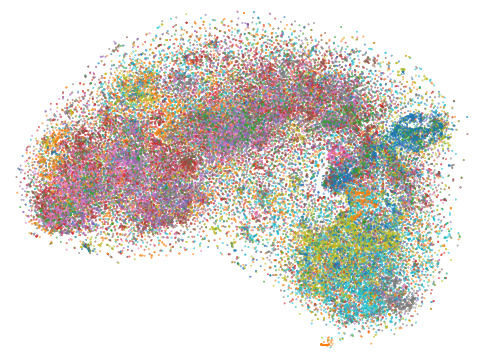}
    \caption{Raw pixel-space embedding of CIFAR-10 computed using t-SNE-CUDA. Notice that while it does have some local continuity, it does not present clear clustering that MNIST has under the L2 metric.}
    \label{fig:cifar_raw}
\end{figure}

While it is natural to expect that the CIFAR-10 dataset is much harder than MNIST due to it's dimensionality, this reasoning lies more in intuition than it does in experimentation. The improved efficiency of t-SNE-CUDA allows us to perform pixel-level experimentation in datasets such as CIFAR-10 \cite{cifar}, whereas previously only embedding-level experiments were possible. This allows us to gain additional insight into the reason of CIFAR-10 being a much harder classification problem than MNIST. Since CIFAR-10 is composed of 32x32x3 images, at a pixel level CIFAR has 50K images at 3072 dimensions. Figure \ref{fig:cifar_raw} shows a t-SNE embedding of the raw pixels CIFAR-10 dataset. We can see immediately from this experiment why classification is much easier on the MNIST dataset. As shown by Figure \ref{fig:vs_fig_mnist}, MNIST has a very clear nearest neighbor structure under the L2 metric in pixel space. In Figure \ref{fig:cifar_raw}, we see that CIFAR does not have the same structure - images that are close in pixel space are likely of many different classes. 

While Figure \ref{fig:cifar_raw} shows that there is clearly some local pixel structure in the dataset, the pixel structure is not as well defined as in the MNIST dataset. Thus, we cannot expect a simple nearest neighbor in the euclidean space to perform well in classification, and we need a non-linear embedding to properly structure the space. Figure \ref{fig:vs_fig_cifar} shows that our non-linear embeddings provides a better L2 structure for our code, making a nearest neighbor classifier in the code-space more efficient (and validating the power of transforming the data with a neural network). 

\subsection{ImageNet}

\begin{figure}
    \includegraphics[width=\linewidth]{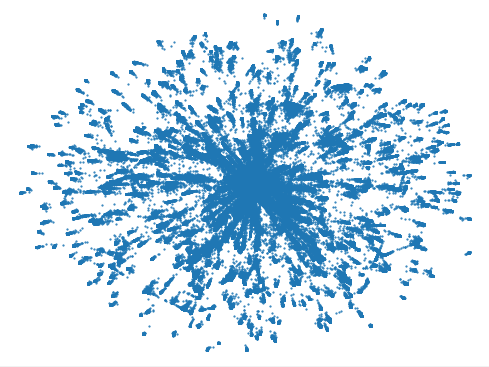}
    \caption{Embedding of the 1.2M VGG16 Codes (4096 dim) computed in 523s. We notice that it is relatively more discrete than the ResNet codes shown in Figure \ref{fig:imagenet}, perhaps suggesting that the classification space is less continuous under the L2 metric. }
    \label{fig:vgg}
\end{figure}

The ImageNet ILSVRC15 dataset \cite{imagenet} is a large-scale image dataset which is particularly popular in computer vision research. ILSVRC15 is composed of 1.2M 224x224x3 full color images. It remains an interesting challenge to explore the ways that different neural networks construct embedding spaces of ILSVRC15. While some previous work has explored codes on the ILSVRC15 validation set \cite{KarImgNetCNN} - such explorations do not provide a full picture of the embedding space of such a large dataset. Figure \ref{fig:vgg} shows the embedding of the VGG19 \cite{simonyan2014very} 4096 dimensional codes for the entire ILSVRC15 dataset, while Figure \ref{fig:imagenet} (Page 1) shows the embeddings using ResNet-200 \cite{he2016deep}. 

An interesting aside is that there are many small, tight clusters in the VGG embedding - each corresponding to a different class. In the ResNet embedding, on the other hand, larger clusters are connected by intermediary data points. We find in general that these inter-connected clusters correspond to coarser classifications such as ``animals" or ``machines." Such more general relationships are not as common in the $L2$ embedding of the VGG codes. These wispy connections suggest that the ResNet embedding space may be more continuous than the VGG embedding space, with points having more inter-class neighbors, while VGG separates classes in a more discrete manner. We can, thus, begin to use the information provided by t-SNE-CUDA to help explore some of the local patterns present in large data/embedding spaces.

\subsection{GloVe}

The GLOVE embedding \cite{pennington2014glove} is a natural language dataset with a vocabulary of over 2.2M words, each embedded in 300 dimensional space. GLOVE is a word-similarity embedding trained on 840B tokens found around the internet. 

\begin{figure}
    \centering
    \includegraphics[width=\linewidth]{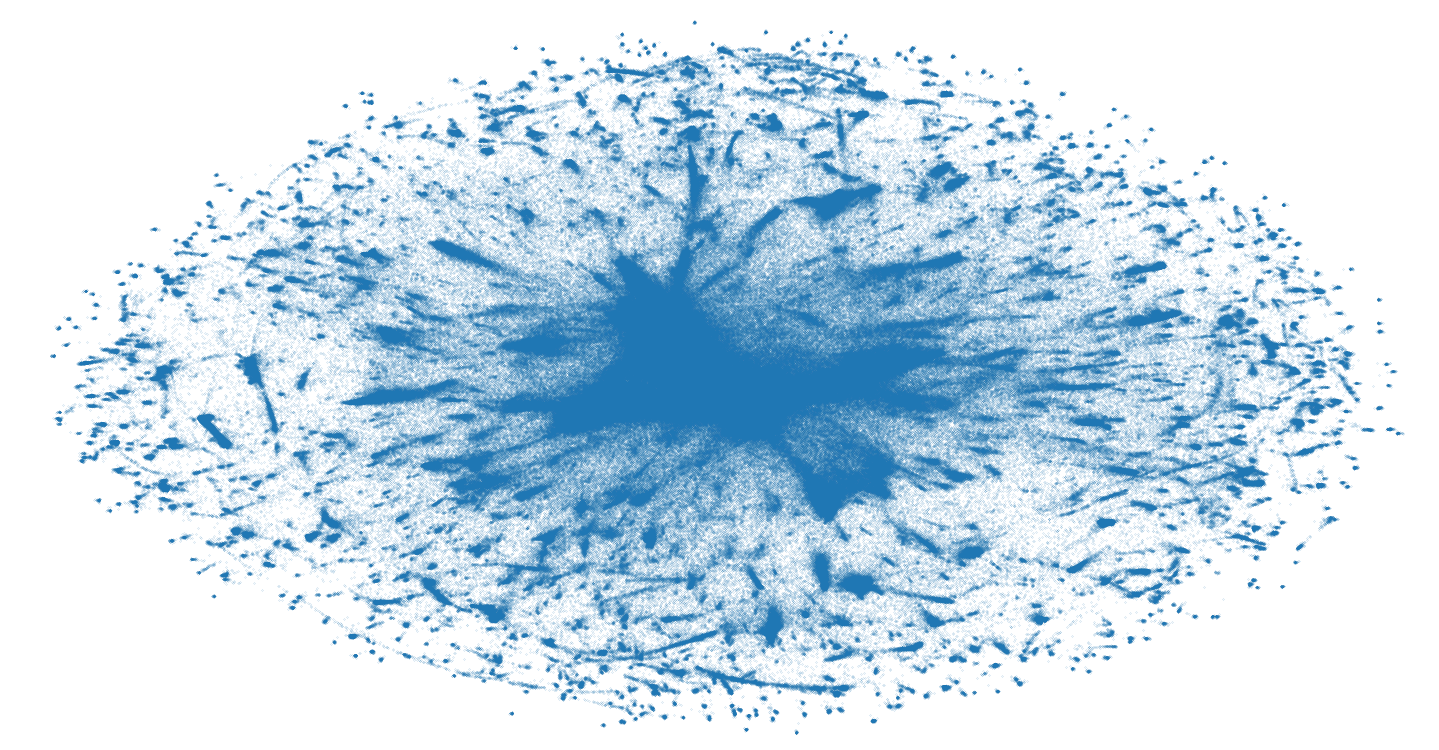}
    \caption{Embedding of the GLOVE space visualized using t-SNE.}
    \label{fig:glove}
\end{figure}

 Figure \ref{fig:glove} shows a coarse plot of the t-SNE that we computed across the \textit{entire} GLOVE vocabulary. An interactive visualization of this dataset is available at \texttt{\url{https://davidmchan.github.io/projects/glove.html}}. Our GLOVE embedding was computed in 573.2s. As far as we know, this is the first time that the entire 2.2M dataset has been visualized. We notice that the L2 metric seems to be a questionable choice for comparing GLOVE vectors. While there are nice clusters of textually similar data (such as french words, dates, and times), semantic clusters seem less prevalent in the embedding space, and clusters appear to be dominated primarily by hamming distance, and not semantic similarity.

\section{Conclusion}
In this paper, we have introduced t-SNE-CUDA, a GPU-accelerated implementation of the t-SNE algorithm. We showed that this algorithm can be optimized by using product-quantization to approximate the nearest neighbours of higher dimensional data points and the Barnes-hut method to approximate gradient computation of t-SNE repulsive forces. With these optimizations, we achieved over 50x speedup over state-of-the-art t-SNE implementations and over 650x over the popular SkLearn library. This speedup enables us to explore previously intractable problems - both in the context of vision (with the ImageNet dataset) and NLP (with the GLoVe embeddings). t-SNE-CUDA is publicly available at \texttt{\url{https://github.com/CannyLab/tsne-cuda}}.

\section*{Acknowledgment} 

The authors would like to thank Dr. James Demmel and Ayd{\i}n Bulu\c{c} for their helpful comments and review when writing this paper. We gratefully acknowledge the support of NVIDIA Corporation with the donation of the Titan X GPU used for this research. We additionally acknowledge the support of the Berkeley Artificial Intelligence Research (BAIR) Lab. 

\bibliographystyle{abbrv}
\bibliography{main}

\vspace{12pt}
\color{red}

\end{document}